\pdfoutput=1

\documentclass[11pt]{article}

\usepackage[]{acl}

\usepackage{times}
\usepackage{latexsym}
\usepackage{multirow}
\usepackage[T1]{fontenc}

\usepackage[utf8]{inputenc}

\usepackage{microtype}

\usepackage{times}
\usepackage{latexsym}

\usepackage{algorithm,amsmath}
\usepackage{pifont}
\usepackage{bbding}
\usepackage{amsfonts}
\usepackage{mathrsfs}
\usepackage[noend]{algorithmic}
\usepackage{url}
\usepackage{makecell}
\usepackage{graphicx}
\usepackage{microtype}
\usepackage{xcolor}
\usepackage{csquotes}
\usepackage{booktabs}

\usepackage{enumitem}
\setlist{nolistsep}

\newcommand{\vspacereduce}{\vspace{-1em}}

\title{REKnow: Enhanced Knowledge for Joint Entity and Relation Extraction}

\author{\textbf{Sheng Zhang}, \ \ \textbf{Patrick Ng}, \ \ \textbf{Zhiguo Wang}, \ \ \textbf{Bing Xiang} \\
  AWS AI Labs \\
  \texttt{\{zshe, patricng, zhiguow, bxiang\} @amazon.com} \\}

\begin{document}
\maketitle
\begin{abstract}
Relation extraction is an important but challenging task that aims to extract all hidden relational facts from the text. 
With the development of deep language models, relation extraction methods have achieved good performance on various benchmarks.
However, we observe two shortcomings of previous methods: first, there is no unified framework that works well under various relation extraction settings; second, effectively utilizing external knowledge as background information is absent.
In this work, we propose a knowledge-enhanced generative model to mitigate these two issues.
Our generative model is a unified framework to sequentially generate relational triplets under various relation extraction settings and explicitly utilizes relevant knowledge from Knowledge Graph (KG) to resolve ambiguities.
Our model achieves superior performance on multiple benchmarks and settings, including WebNLG, NYT10, and TACRED.

\end{abstract}

\section{Introduction}

Numerous relational facts are hidden in the emerging corpus. The extraction of relational facts is beneficial to a diverse set of downstream problems. 
For example, relational facts facilitate the development of dialogue systems with lower labor cost \cite{wen2016network, dhingra2016towards}. Knowledge Graph (KG) construction leverages relational fact extraction of everyday news to keep its KGs up-to-date \cite{shi2018open}. Even in the field of e-commerce, relation extraction of product descriptions helps build category information for products \cite{dong2020autoknow}.

\begin{figure}[ht]

\resizebox{1\columnwidth}{!}{%
 \centering
  \includegraphics[width=0.9\textwidth]{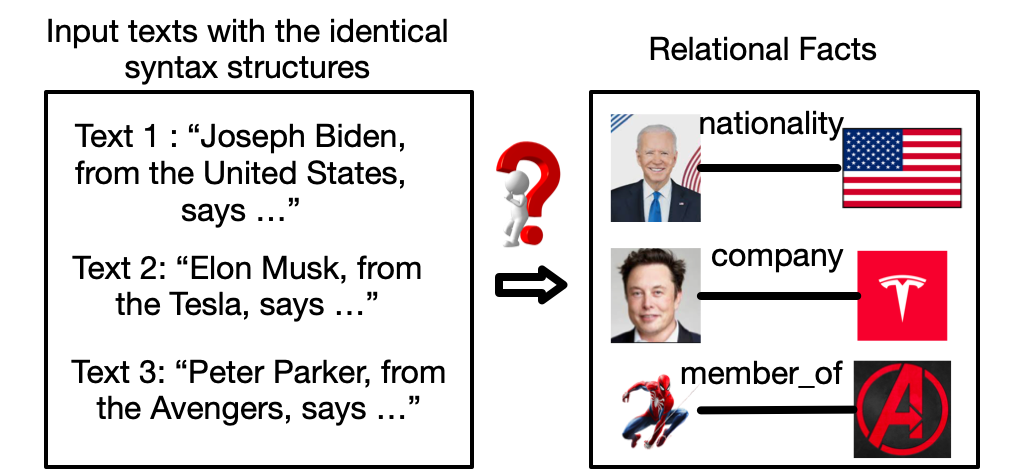}
  }
 \caption{The three examples have identical syntax structures, but their corresponding relational facts
 are different. Therefore, background knowledge is very crucial to correctly extract relational facts.
}
 \label{fig:0}
\vspacereduce
\end{figure}

\begin{table}
\resizebox{1\columnwidth}{!}{%
\begin{tabular}{@{}lp{7.5cm}@{}}
\toprule
Task name                           & \textbf{Entity Type aware Relation Classification}                                                                                                                                                                                       \\ \cmidrule(l){2-2} 
Text $\dagger$  & Born in London in 1939 the son of a Greek   tycoon, Negroponte grew up in Britain, Switzerland and the United   States                                                                                                                                                                 \\ \cmidrule(l){2-2} 
Entity Positions                   &  Negroponte: (51, 61) \\ 
                                 & United States: (103, 116) \\
Entity Types                                    &  Negroponte: PERSON \\ 
                                & United States: LOCATION  \\ 
                                    \midrule
Task   name                         & \textbf{Relation Classification}                                                                                                                                                                                \\ \cmidrule(l){2-2} 

Text $\ddagger$ & The   cutting machine contains 13 circular blades mounted on a cutting axis.                                                                                                                                                                                                            \\ \cmidrule(l){2-2} 
Entity Positions                                    & machine: (12, 19)   \\
                                                    & blades: (41, 47)                                                                                                         \\ \midrule
Task   name                         & \textbf{Joint Entity and Relation Extraction}                                                                                                                                                                                                       \\ \cmidrule(l){2-2} 
Text $\mathsection$    & It is Japan's second-biggest automaker,   behind Toyota and ahead of Nissan.                   
\\ \bottomrule
\end{tabular}}
\caption{Different type of relational fact extraction tasks. $\dagger$  example is obtained from TACRED \citep{zhang2017tacred}; $\ddagger$ from SemEval2010 \cite{hendrickx2019semeval}; $\mathsection$ from NYT10 \cite{riedel2010modeling}.  
See Section \ref{sec:problem_def} for more details.} 
\label{table:1}
\vspacereduce
\end{table}


Although relation extraction (RE) has been widely studied, we observe two shortcomings of previous work.
First, there is no unified framework for different settings of relation extraction tasks. 
Table \ref{table:1} summarizes the popular relational fact extraction tasks.
The formats of input and output vary by task.
In the Entity Type aware Relation Classification task, models take the plain text and entity positions and entity types as input and predict relation types for all entity pairs.
For the Relation Classification task, entity types are not provided.
In the Joint Entity and Relation Extraction task, models only take plain texts as input and predict all possible entities and relations among them. 
Previous work usually designs different model frameworks to handle these tasks separately \citep{joshi2020spanbert, yamada2020luke, soares2019matching, wei2019novel, cabot2021rebel}.  
Also, directly applying Joint Entity and Relation Extraction models to Relation Classification tasks would lead to a bad performance due to the lack of fully usage of entity information compared with task-specific models.

Second, conventional relation extraction models do not utilize external knowledge to resolve relational ambiguities.
However, background knowledge of entities can be very crucial for models to figure out the right relations.
For the three examples in Figure \ref{fig:0}, 
they have the same syntax structure: "\texttt{[entity1] from the [entity2], says}".
The main information for models to predict different relation types for them is the background knowledge of these involved entities.
For example, the facts of "Joseph Biden is a human" and "the United States is a country" determine the "nationality" relation. The facts of "Elon Musk is a human" and "Tesla is an enterprise" determine the "company" relation. The facts of "Peter Parker is a fictional human" and "Avengers is a fictional organization" determine the "member\_of" relation.
Lack of the background knowledge, it is very difficult for models to disambiguate.

Our solution to the first issue is to utilize a generative model framework.
The popular sequence-to-sequence frameworks are naturally equipped to solve multiple tasks with variations in input and output formats. 
Moreover, previous studies \citet{ye2020contrastive} and \citet{josifoski2021genie} have shown the effectiveness of the generative framework for each specific RE task.
In this work, we aim to unify all types of RE tasks with a generative model.
To address the second issue, we propose to leverage the well-organized information from the knowledge graph (KG).
We employ an entity linking model to connect all entities within the plain text with their corresponding entries in KG, then explicitly feed the knowledge of these entities into the RE model for relation prediction.
Previous methods \citep{zhang2019ernie, wang2020k, liu2020k} have shown that injecting KG into models during pre-training can boost the performance in downstream tasks.
However, involving KG during pre-training can be very expensive, and the model may suffer the catastrophic forgetting issue.
Differently, our method directly feeds relevant knowledge into the downstream model, thus can be more efficient and effective.

In this paper, we propose a unified generative framework for \textbf{R}elation \textbf{E}xtraction with \textbf{Know}ledge enhancement (\textbf{REKnow}). Our generative framework is trained to sequentially generate relational facts token-by-token. 
Also, our framework enhances the input plain text with relevant KG information.  
Our experiments show that the proposed framework significantly boosts the extraction performance on multiple benchmarks and task settings. 
To our best knowledge, this is the first work to integrate KG into a generative framework for relational fact extraction.
In summary, our contributions are threefold: 
\begin{itemize}
    \item We construct a novel knowledge-enhanced generative framework for relational fact extraction task.
    \item We provide KG grounding predictions for public relation extraction benchmark datasets.\footnote{Code, models and grounded KG will be available at GitHub.}
    \item Our generative model achieves superior performance on multiple benchmarks and settings, including ACE2005, NYT10, and TACRED.
\end{itemize}

\section{Problem Definition } \label{sec:problem_def}

\begin{figure*}[ht]
  \centering
  \includegraphics[width=0.9\textwidth]{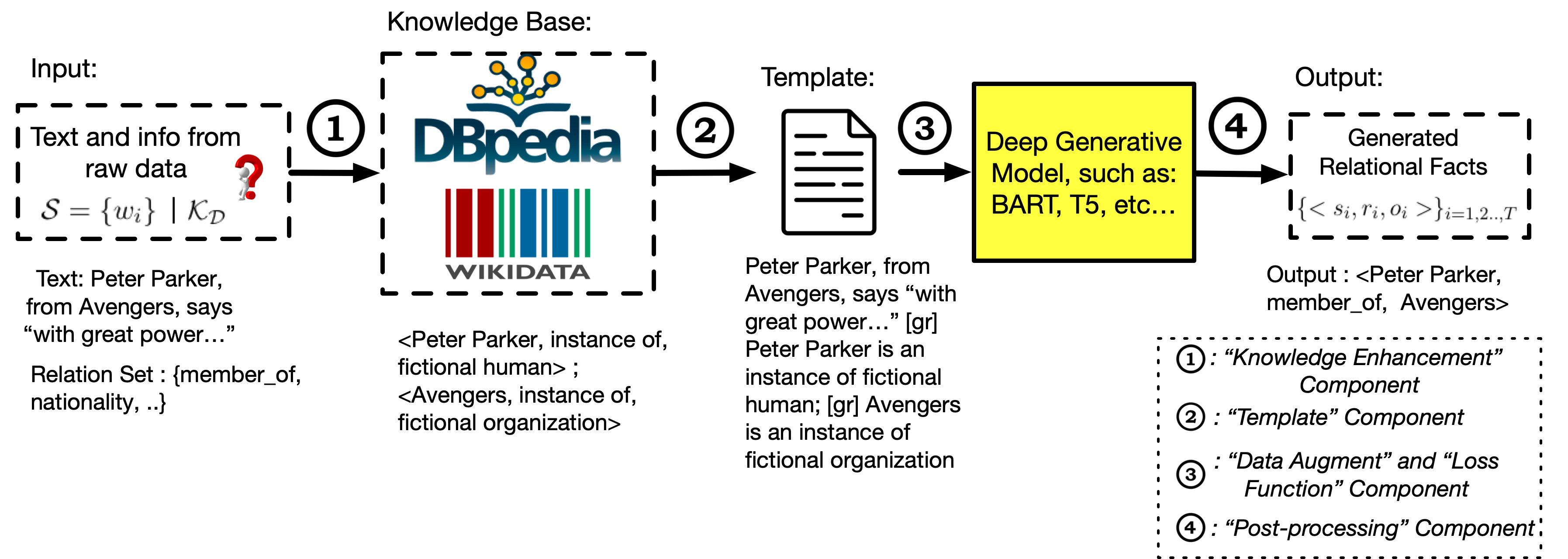}
 \caption{Overall REKnow framework. See Section \ref{sec:model} for more details about each component.
}
 \label{fig:1}
\vspacereduce
\end{figure*}

All tasks in Table \ref{table:1} are formally formulated as follows.
We denote the plain text as $\mathcal{S} = \{w_i\}_{i = 1,2,..,N}$, where $w_i$ represents a word, and $N$ is the text length. 
The knowledge provided in a dataset $\mathcal{D}$ is denoted as $\mathcal{K}_{\mathcal{D}}$, such as pre-defined relation set $\mathcal{R}$, entity positions $\mathcal{E}_{pos}$, and entity types $\mathcal{E}_{type}$. 
The goal is to find all relation triples $\{<\mathbf{s}_i,\mathbf{r}_i,\mathbf{o}_i>\}_{i = 1,2..,T}$ in text $\mathcal{S}$, where 
$\mathbf{s}_i$ and $\mathbf{o}_i$ are subject and object entities in $\mathcal{S}$, 
$\mathbf{r}_i$ belonging to $\mathcal{R}$ is the relation between subject and object, and $T$ is the number of relation triples.
The RE problem can be formulated as follows:
\begin{equation}\label{eq:general_format}
   f(\mathcal{S} | \mathcal{K}_{\mathcal{D}}) = \{<\mathbf{s}_i,\mathbf{r}_i,\mathbf{o}_i>\}_{i = 1,2..,T}. 
\end{equation}

For different tasks, $\mathcal{K}_{\mathcal{D}}$ may contain a different set of knowledge. 
We consider three tasks: (1)~Entity Type aware Relation Classification, where $\mathcal{K}^{ETRC}_{\mathcal{D}} = \{\mathcal{R}, \mathcal{E}_{pos}, \mathcal{E}_{type}\}$; (2)~Relation Classification, where $\mathcal{K}^{RC}_{\mathcal{D}} = \{\mathcal{R}, \mathcal{E}_{pos}\}$; and (3)~Joint Relation and Entity Extraction $\mathcal{K}^{JREE}_{\mathcal{D}} = \{\mathcal{R}\}$. 
We denote tasks (1)~and (2)~as entity position-aware case and task (3)~as entity position-absent case for simplicity in the following sections.

\section{Relation Extraction with Knowledge-Enhanced Generative Model}\label{sec:model}

\paragraph{Overview}

Figure \ref{fig:1} shows the workflow of our method. 
The relational fact extraction is modeled as a seq2seq generative model, where knowledge-enhanced text is passed to the encoder, and the relational facts are generated sequentially from the decoder.
We use text $\mathcal{S} = $\textit{``Peter Parker, from Avengers, says \ldots''} and relation set $\mathcal{R} = $  \{\textit{member\_of}, \textit{nationality}, \ldots\} as a running example.
First, the plain text $\mathcal{S}$ is grounded to the public Knowledge Base (KB) Wikidata\footnote{\url{https://www.wikidata.org/}} to obtain relevant background knowledge $\mathcal{K}_{E}$, where the subscript $E$ refers to external knowledge. 
In this example, we ground \textit{``Peter Parker''} to \url{https://www.wikidata.org/wiki/Q23991129} and \textit{``Avengers''} to \url{https://www.wikidata.org/wiki/Q322646}.
We use the ``instance of'' property in Wikidata to retrieve the background knowledge, e.g. <\textit{Peter Parker}, instance of, \textit{fictional human}> and <\textit{Avengers}, instance of, \textit{fictional organization}>.
Then, we combine the plain text $\mathcal{S}$, knowledge from the dataset $\mathcal{K}_{\mathcal{D}}$, and external knowledge $\mathcal{K}_{E}$ via a pre-defined template $\mathcal{T}$. 
Finally, the combined text is passed into a generative language model, such as BART \cite{lewis2019bart} or T5 \cite{raffel2019exploring}, and the model directly generates the target relational triples token by token, e.g. \textit{``Peter Parker member\_of Avengers''}. 



\paragraph{Knowledge-Enhancement} We leverage an entity linking model to ground texts to knowledge bases. 
In the Relation Classification task, the span of an entity $\mathcal{E}_{pos}$ is given. 
Therefore, we apply the bi-encoder entity linking method (BLINK) for entity linking \cite{wu2019scalable}. 
BLINK uses a two-stage method that links entities to Wikipedia, based on fine-tuned BERT architectures. 
First, BLINK performs the retrieval step in a dense space defined by a bi-encoder that independently embeds the entity $\mathbf{e} \in \mathcal{E}_{pos}$ and the text $\mathcal{S}$.
In the second stage, each candidate in the KG is examined with a cross-encoder ranking model by concatenating the entity and the text.
An approximate nearest neighbor search method FAISS \cite{johnson2019billion} is applied to search the large-scale KG efficiently.
In the task of Joint Entity and Relation Extraction, the span of an entity is absent. We apply an efficient one-pass end-to-end Entity Linking method (ELQ) \cite{li-etal-2020-efficient}. 
ELQ follows a similar architecture as BLINK, but the bi-encoder jointly performs mention detection and linking in one pass. 

After we obtain the grounded Wikipedia entries for entities in the text, we use the public-source pywikibot\footnote{https://github.com/wikimedia/pywikibot} to retrieve the entity knowledge from Wikidata, i.e. the knowledge provided by the "instance of" property.
If multiple "instances of" properties exist, we select the one with the highest frequency over the training data.
Formally, for the entity position-aware case, entity linking is formulated as:
\begin{equation} \label{eq:el1}
    \mathcal{K}_E = \{\mathcal{E}_{type}\} = EL_{BLINK}(\mathcal{S} | \{\mathcal{E}_{pos}\}).
\end{equation}
The entity position-absent case is formulated as:
\begin{equation} \label{eq:el2}
    \mathcal{K}_E = \{\mathcal{E}_{pos}, \mathcal{E}_{type}\} = EL_{ELQ}(\mathcal{S}).
\end{equation}
More implementation details will be discussed in Section \ref{subsec:model_implement}.

\paragraph{Template} We consider two templates to incorporate entity-related knowledge for both entity position-aware and entity position-absent cases in our application.
For text $\mathcal{S}$, we denote the joint knowledge from dataset and external knowledge as $\mathcal{K}_\mathcal{S} = \mathcal{K}_{\mathcal{D}} \bigcup \mathcal{K}_E$. 
For the entity position-aware case, we construct the template as 
\begin{equation*}
    \begin{split}
        & \mathcal{T}_1(\mathcal{S} | \mathcal{K}_\mathcal{S}) =\\ & \{w_1, .., w_{i-1}, [es], w^{s_t}_{i},.., w^{s_t}_{j}, [gr], \mathbf{t}_{s_t}, w_{j + 1}, .., \\
        &  w_{k - 1} , [es], w^{o_t}_{k},.., w^{o_t}_{l}, [gr], \mathbf{t}_{o_t}, w_{l + 1}, .., w_{N}\},
    \end{split}
\end{equation*}
where $[es]$ and $[gr]$ are special tokens to denote entity mention start and grounded entity information position, respectively. In the above case, the token span for subject $\mathbf{s}_t$ ranges from $i$-th to $j$-th token in original text, similar notation is applied to object $\mathbf{o}_t$.
Entity type for entity $\mathbf{e}$ is denoted as $\mathbf{t}_{e}$.

For the entity position-absent case, we construct the template by concatenating the entity knowledge information at the end of the original text as follows,
\begin{equation*}
    \begin{split}
        & \mathcal{T}_2(\mathcal{S} | \mathcal{K}_\mathcal{S}) =\\ & \{w_1, w_2, .., w_{N}, [gr], \mathbf{e_1} is \ an \  instance \ of \ \mathbf{t}_{e_1},  \\ 
        & ... , [gr], \mathbf{e}_M \ is \ an \ instance \ of \ \mathbf{t}_{e_M} ..\},
    \end{split}
\end{equation*}
where $\mathbf{e}_i, i = 1, 2,.., M$ are the retrieved entities via the entity linking system, 
and $\mathbf{t}_{e_i}$ are corresponding entity knowledge from public  KB\footnote{In the T5 paper, authors added a prefix to distinguish different tasks during the pre-training stage.
Therefore, if T5 pre-trained model is adopted, we add the prefix ``\texttt{summary:}" in the template for our fine-tuning stage.}.

\paragraph{Data Augmentation}
If multiple triples exist in a single input text, the order of target relational triples may affect the performance in the seq2seq model, which is known as \textit{exposure bias} \cite{ranzato2015sequence}.
In previous relation extraction work, researchers attempted to alleviate exposure bias by replacing maximum likelihood estimation (MLE) loss with reinforcement learning-based loss \cite{zeng-etal-2019-learning} or unordered multi-tree structure-based loss \cite{zhang2020minimize}.
In our work, we adapt the data augmentation strategy by shuffling the order of target relational triples and adding the shuffled triples to the original dataset during the training phase.
Our ablation study in Section \ref{subsec:result} shows that the data augmentation strategy can significantly improve the performance.

\paragraph{Loss Function}
We apply a generative model, such as BART \cite{lewis2019bart} and T5 \cite{raffel2019exploring}, to input text $\mathbf{x} = \mathcal{T}(\mathcal{S}| \mathcal{K}_\mathcal{S})$. 
The output of triples $\mathbf{y} = \{ \mathbf{s}_1\ \mathbf{r}_1\ \mathbf{o}_1; \mathbf{s}_2\ \mathbf{r}_2\ \mathbf{o}_2;  ... ;\mathbf{s}_T\ \mathbf{r}_T\ \mathbf{o}_T\}$ is generated in the text format in a single pass, where token ``;'' is used to separate different triples in the output.
The generative model needs to model the conditional distribution $p(\mathbf{y} | \mathbf{x})$. 
The output tokens come from a fixed size vocabulary $V$.
The distribution $p(\mathbf{y} | \mathbf{x})$ can be represented by the conditional probability of the next token given the previous tokens and input text:
$$
p(\mathbf{y} | \mathbf{x}) = \prod_{i = 1}^{|y|} p(y_i | y_{<i}, \mathbf{x}),
$$
where $y_i$ denotes the $i$-th token in $\mathbf{y}$. 
The subject $\mathbf{s}_t$ and object $\mathbf{o}_t$ are tokens shown in the text $\mathcal{S}$, and $\mathbf{r}_t$ is a relation from the relation set $\mathcal{R}$, i.e. $\mathbf{s}_t, \mathbf{o}_t \subset \mathcal{S}$, $\mathbf{r}_t \in \mathcal{R}$ for $t = 1, 2,..,T$.
For training set $D = \{(\mathbf{x}^i, \mathbf{y}^i)_{i = 1, 2.., |D|}\}$, we train the generative model with parameter $\theta$ using MLE by
minimizing the negative log-likelihood over $D$: 
$$
\mathcal{L}(D) = -\sum_{i = 1}^{|D|}\log p_\theta(\mathbf{y}^i | \mathbf{x}^i).
$$

\paragraph{Post-processing}

The generated output $\mathbf{y}$ is in the text format, which should be transformed to standard triples format $\{<\mathbf{s}_i,\mathbf{r}_i,\mathbf{o}_i>\}_{i = 1,2..,T}$ for evaluation.
Therefore, we apply some post-processing steps, including \textit{split}, \textit{delete} and \textit{replace}, to transform the generated output.
Token ``;'' is used to separate triples in the training stage, hence the generated output can be \textit{split} by token ``;'' to obtain different triples in text format.
For each generated triple, the relation should belong to the relation set $\mathcal{R}$.
If no valid relation is found in the generated triple, we will \textit{delete} the corresponding triple. 
In Relation Classification task, since the gold entities $\mathcal{E}$ are given, we match the generated entities to the gold entities using Levenshtein similarity \citep{navarro2001guided}. In the Relation Classification task, we would \textit{delete} the generated triple if the generated entities show low similarity to all golden entities,
i.e. $\max_{\mathbf{e}_i \in \mathcal{E}} L_{sim}( \hat{\mathbf{e}}, \mathbf{e}_i) < \epsilon$ 
where $L_{sim}$ denotes Levenshtein similarity, $\hat{\mathbf{e}}$ is the generated entity, and $\epsilon$ is a threshold\footnote{We use $\epsilon = 0.85$ in our experiments}.
Otherwise, we \textit{replace} the generated entity with the gold entity with the highest Levenshtein similarity.
In the Joint Entity and Relation Classification task, the generated entity should be shown in the original text. 
Therefore, we match the generated entities to the set of sub-spans in the text, which is motivated by span-level prediction in Name Entity Recognition (NER) research \cite{li2019unified, yu2020named, fu2021spanner}.
We \textit{replace} a generated entity with the sub-span of text in the text with the highest Levenshtein similarity. 

\section{Related Work}\label{sec:related}

\paragraph{Relational Fact Extraction}

We group the relational fact extraction tasks with pre-defined relation set into three categories based on different levels of information provided in the data and introduce the related methods accordingly.
In the first category, for data with entity span as well as corresponding entity types, SpanBERT \cite{joshi2020spanbert} built the relation classification model by replacing the subject and object entities with their NER tags such as \textit{``[CLS][SUBJ-PER] was born in [OBJ-LOC], Michigan, \ldots''}. 
A linear classifier was added on top of the [CLS] token to predict the relation type in their applications.
LUKE \cite{yamada2020luke} constructed an additional Entity Type Embedding layer in their model construction to utilize the entity type information.

In the second category, data come with an entity span.
It is the most popular and standard format in the RE task. 
Traditional methods \cite{kambhatla2004combining, zhou2005exploring} extracted the relation between entities based on feature engineering, which incorporated semantic, syntactic, or lexical features in the text.
After the fast development of the deep language model, like BERT \cite{devlin2018bert}, BERT-Entity \cite{soares2019matching} utilized the entity span information by adding a mention pooling layer on top of the BERT model. 

In the third category, only text is provided in the data. 
In such a case, models were developed to extract both entity and relation.
For pipeline-based methods, 
NER techniques \cite{chiu2016named, huang2015bidirectional, fu2021spanner} were used to detect the span of entities, then relation classification models are applied based on the detected entities.
Researchers \cite{li2014incremental, miwa2014modeling} also noticed that sharing parameters between NER and RE models are important in the pipeline model.
For example, Pure \cite{zhong2020frustratingly} built separate encoders for entities and relations, where the side product of the entity model is fed into the relation model as input.
For joint extraction methods, Copy-RE \cite{zeng2018extracting} built the relation extraction model based on the copy mechanism. Tplinker \cite{wang2020tplinker} jointly extracted the relation and entity in novel handshaking tagging schema, which are constructed to deal with the issue of overlapping entities.
In this work, we propose a unified framework for the relation extraction task for all cases above. 
Compared to a unified framework for NER \cite{yan2021unified} tasks, our work focuses on relational triplet extraction.

\paragraph{Knowledge-Enhanced Information Extraction}

With the development of KG construction, many NLP tasks benefit from the external well-organized knowledge in KG. 
We group knowledge-enhanced methods in the RE task into two categories based on the knowledge-ingestion stage. 
The first type is to use the KG in the pre-training stage. The key challenges for such knowledge-ingestion type are Heterogeneous Embedding Space (HES), which means the embedding spaces for KG and text would be heterogeneous, and Knowledge Noise (KN), which represents information in KG would divert the text from its correct meaning \cite{liu2020k}. 
ERNIE \cite{zhang2019ernie} solved the HES by proposing a pre-training task that requires the model to predict entities based on the given entity sequence rather than all entities in KGs.
K-Adapter \cite{wang2020k} constructed a neural adapter to infuse different kinds of knowledge.
REBEL \cite{cabot2021rebel} used BART-large as the base model to generate relation triplets, which is pretrained on Wikidata. However, since the method was trained on the Wikidata with 220 different relation types, it may introduce knowledge noise implicitly.
MIUK \cite{li2021multi} leveraged the uncertainty characteristic in ProBase \cite{wu2012probase} across three views: mention, entity and concept view. 
Numerical studies about integrating knowledge into NLU tasks \cite{xu2021does} showed that KG information can increase the performance by a significant amount.

Another type is to use the KG in downstream tasks directly. 
One of the popular implementations is distantly supervised relation extractors \citealp{mintz2009distant}, which leveraged the information in Freebase \cite{bollacker2008freebase} to extract relations.
\citealp{liu2020knowledge} utilized KG to conduct information extraction via RL framework.
Our method uses entities linking to automatically ground entities in the text to public KG. 
The experiment results show that the entity type in KG helps boost the performance in most cases.

\section{Experiments}\label{sec:exp}
In this section, we compare our proposed method to the state-of-the-art of relational fact extraction methods --- demonstrating its effectiveness and ability to extract relational facts under multiple tasks.
Our codes will be made publicly available.

\subsection{Data}

\begin{table}[]
\resizebox{1\columnwidth}{!}{%
\tiny
\begin{tabular}{@{}lll@{}}
\toprule
\multirow{2}{*}{Dataset} & Data Size          & Ave. Triple Size \\
                         & (train/val/test)   & (train/val/test) \\ \midrule
TACRED                   & 10,021/3,894/2,307 & 1.30/1.40/1.44   \\
SemEval2010              & 6,507/1,493/2,717  & 1.00/1.00/1.00      \\
NYT10                    & 56,196/5,000/5,000 & 2.01/2.02/2.03   \\
Webnlg                   & 5,019/500/703      & 2.74/3.11/2.82   \\
ACE2005                  & 2,619/648/590      & 1.83/1.82/1.95   \\ \bottomrule
\end{tabular}
}
\caption{Statistics of relational fact extraction benchmark datasets. Avg Triple Size represents the average size of triples in each input data. }
\label{table:2}
\end{table}

Several public available RE datasets are evaluated in our numerical experiments.
\textbf{TACRED} \citep{zhang2017tacred} is a large supervised RE dataset,
which is derived from the TAC-KBP relation set, with labels obtained via crowdsourcing.
It contains 41 valid relation types as well as one null relation type and 12 entity types. 
Although alternate versions of TACRED have been published recently \cite{alt2020tacred, stoica2021re}, we use the original version of TACRED, with which the state of the art is mainly tested.
\textbf{SemEval 2010 Task 8} \citep{hendrickx2019semeval} is a benchmark dataset for evaluating classification of semantic relations between pairs of nominal entities, such as ``part-whole'', ``cause-effect'', etc.
It defines 9 valid relation types as well as one ``other" relation type and no entity type is provided.  
We follow the split setting as in \cite{soares2019matching} for numerical study.
\textbf{NYT10} dataset \citep{riedel2010modeling} was originally produced by distant supervision method from 1987-2007 New York Times news articles. 
Original data are adapted by \cite{zeng2018extracting} for complicated relational triple extraction task, where EntityPairOverlap (EPO) and SingleEntityOverlap (SEO) widely existed in the dataset.
\textbf{WebNLG} dataset \citep{gardent2017creating} was originally created for Natural Language Generation task. 
\citealp{zeng2018extracting} adapted WebNLG for relational triple extraction task as well.
In our experiment, we evaluate the proposed method on NYT10 and WebNLG datasets with whole entity span as in prior works \cite{ wang2020tplinker, ye2020contrastive} other than datasets with the last word of the entities \cite{zeng2018extracting, wei2019novel}.
\textbf{ACE2005} \cite{ace2005} is developed by the Linguistic Data Consortium (LDC) containing a collection of documents from a variety of domains including news and online forums. 
In the dataset, 6 relation types between the entities are provided.
In our implementation, we focus on the RE task in ACE2005, hence we only keep the sentences with valid triples.
The detailed statistics about all datasets are summarized in Table \ref{table:2}. 
For the evaluation, we report the standard micro Precision, Recall, and F1-score.

\begin{table}[]
\resizebox{1\columnwidth}{!}{%
\begin{tabular}{@{}p{1.6cm}llll@{}}
\toprule
Dataset                      & Method                                  & Prec. & Rec. & F1            \\ \midrule
\multirow{8}{*}{TACRED}      & ERNIE \cite{zhang2019ernie}             & 80.0  & 66.1 & 68.0          \\
                             & SpanBERT \cite{joshi2020spanbert}       & 70.8  & 70.9 & 70.8          \\
                             & K-Adapter \cite{wang2020k}              & 68.9  & 75.4 & 72.0          \\
                             & RoBERTa \cite{wang2020k}                & 70.2  & 72.4 & 71.3          \\
                             & LUKE \cite{yamada2020luke}              & 70.4  & 75.1 & 72.7 \\
                             & RECENT \cite{Lyu2021RelationCW}              & 90.9  & 64.2 & \textbf{75.2} \\
                             \cmidrule(l){2-5} 
                             & REKnow                           & 76.2  & 74.1 & \textbf{75.1} \\
                             &                                  & (0.51)  & (0.45) & (0.47) \\
                              \midrule
\multirow{5}{*}{Semeval} & CR-CNN    \cite{santos2015classifying}  & -     & -    & 84.1          \\
                             & BERT-Entity   \cite{soares2019matching} & -     & -    & 89.2          \\
                             & BERT-MTB \cite{soares2019matching}      & -     & -    & \textbf{89.5} \\ \cmidrule(l){2-5} 
                             & REKnow                            & 88.1  & 91.4 & \textbf{89.8} \\
                             &                                   & (0.30)  & (0.59) & (0.26)        \\ \bottomrule
\end{tabular}
}
\caption{Main result for entity position-aware case. The top two performing models per dataset are marked in bold. In REKnow, the pre-trained model is T5-large and entity type is obtained by Equation \eqref{eq:el1}. The results are averaged over 5 seeds for REKnow. Standard deviations are indicated within parentheses.}
\label{table:main_1}
\end{table}


\subsection{Model Implementation Details} \label{subsec:model_implement}
All experiments are conducted with 8-cores NVIDIA Tesla V100 GPUs with 2.5 GHz (base) and 3.1 GHz (sustained all-core turbo) Intel Xeon 8175M processors.
\paragraph{Decoder} For the BART model, \citealp{shakeri2020end} found that using a variant of nucleus sampling \citep{holtzman2019curious} can increase the diversity of generated output compared to beam search.
We observe similar findings in our preliminary experiments; therefore, we adopt the \textit{Topk+Nucleus} decoding strategy in our decoding step for the BART model. 
To be specific, we pick top $k = 20$ tokens, and within top k, tokens with top 95\% probability mass are picked in each generated step. 
For the T5 model, we simply adopt the greedy decoding strategy, which can generate triples well.
\paragraph{Entity Linking} We use the pre-trained BLINK and ELQ model\footnote{\url{https://github.com/facebookresearch/BLINK} (MIT License)} to align Wikipedia entity for entity span aware and absent cases, respectively.
The entity linking models are trained on 5.9M entities from May 2019 English Wikipedia dump.
We follow the same threshold score (-4.5) as in the original ELQ implementation, and the top 1 grounded Wikipedia entity is used.
For dataset Semeval 2010, the entities are nominal, hence we use the property ``subclass of" 
of the retrieved entity as entity types.
For datasets TACRED, NYT10, WebNLG and ACE2005, we use property ``instance of" 
to obtain corresponding entity types.

\subsection{Results}\label{subsec:result}

\begin{table}[]
\resizebox{1\columnwidth}{!}{%
\begin{tabular}{@{}p{1.2cm}llll@{}}
\toprule
Dataset                  & Method                                        & Precison & Recall & F1            \\ \midrule
\multirow{11}{*}{NYT}    & NovelTagging \cite{zheng2017joint}            & 32.8     & 30.6   & 31.7          \\
                         & MultiHead   \cite{bekoulis2018joint}          & 60.7     & 58.6   & 59.6          \\
                         & ETL-Sapn \cite{yu2019joint}                   & 85.5     & 71.7   & 78.0          \\
                         & Tplinker \cite{wang2020tplinker}         & 91.4     & 92.6   &     92.0 \\
                         & CopyRE* $\dagger$  \citep{zeng2018extracting} & 61       & 56.6   & 58.7          \\
                         & CopyMTL* $\dagger$  \cite{zeng2020copymtl}    & 75.7     & 68.7   & 72.0          \\
                         & TANL \cite{paolini2021structured} $\dagger$  &  -   & -  & 90.8         \\
                         & REBEL \cite{cabot2021rebel} $\dagger$                           & -        & -      & \textbf{93.4} \\
                         & CGT $\dagger$  \citep{ye2020contrastive}      & 94.7     & 84.2   & 89.1          \\ \cmidrule(l){2-5} 
                         & REKnow                                  & 93.1     & 94.1   & \textbf{93.6} \\
                         &                                         & (0.18)   & (0.17)  & (0.17)      \\ \midrule
\multirow{9}{*}{Webnlg}  & NovelTagging    \cite{zheng2017joint}         & 52.5     & 19.3   & 28.3          \\
                         & MultiHead   \cite{bekoulis2018joint}          & 57.5     & 54.1   & 55.7          \\
                         & ETL-Sapn \cite{yu2019joint}                   & 84.3     & 82.0   & 83.1          \\
                         & Tplinker \cite{wang2020tplinker}         & 88.9     & 84.5   & \textbf{86.7} \\
                         & CopyRE* $\dagger$  \citep{zeng2018extracting}          & 37.7     & 36.4   & 37.1          \\
                         & CopyMTL*  $\dagger$ \cite{zeng2020copymtl}             & 58.0     & 54.9   & 56.4          \\
                         & CGT $\dagger$ \citep{ye2020contrastive}                 & 92.9     & 75.6   & 83.4          \\ \cmidrule(l){2-5} 
                         & REKnow                                 & 90.4  & 87.9   & \textbf{89.1} \\
                         &                                               & (0.17)     & (0.33)   & (0.24) \\
                         
                         \midrule
\multirow{7}{*}{ACE2005} & Attention \cite{katiyar2017going}             & -        & -      & 55.9          \\
                         & DYGIE \cite{luan2019general}                  & -        & -      & 63.2          \\
                         & DYGIE++ \cite{wadden2019entity}               & -        & -      & 63.4          \\
                         & Pure-Bb \cite{zhong2020frustratingly}     & -        & -      & 66.7          \\
                         & Pure-Alb \cite{zhong2020frustratingly}      & -        & -      & \textbf{69.0} \\ \cmidrule(l){2-5} 
                         & REKnow                              & 71.3      & 67.6     & \textbf{69.4} \\
                         &                     & (0.81)   & (0.30)   & (0.54)          \\ \bottomrule
\end{tabular}
}
\caption{Main result for entity position-absent case. * denotes the methods that only evaluate the last token of the entity. $\dagger$ denotes the competitors are generative-based models. The top two performing models per dataset are marked in bold. In REKnow, the pre-trained model is T5-large and entity type is obtained by Equation \eqref{eq:el2}. The results are averaged over 5 seeds for REKnow. Standard deviations are indicated within parentheses.}
\label{table:main_2}
\end{table}
\begin{figure}[ht]
\centering
\resizebox{0.9\columnwidth}{!}{%
  \includegraphics[width=1\textwidth]{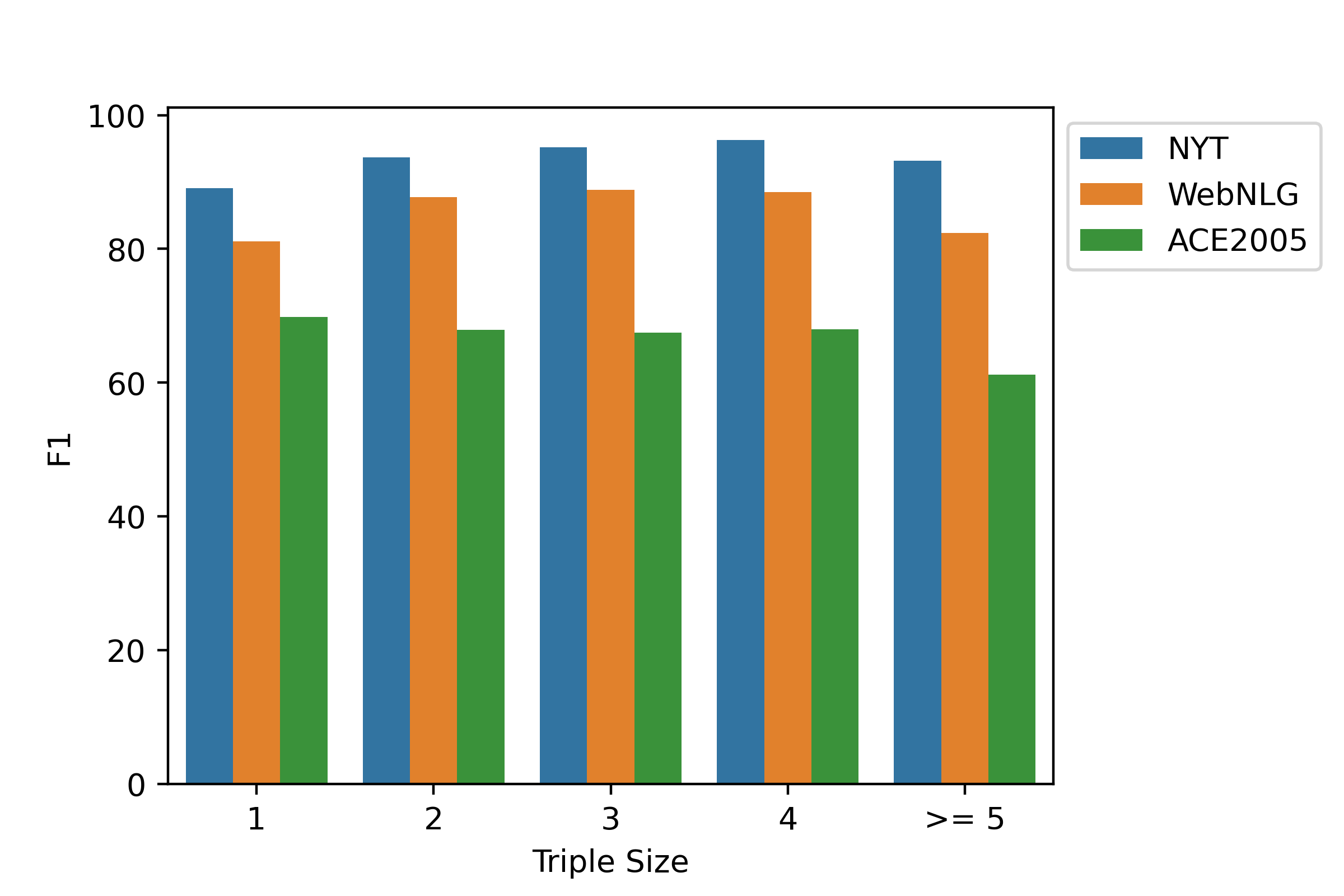}
  }
 \caption{F1 Score under different triple sizes for REKnow. The \textit{triple size} is the number of triples hidden in a sentence. 
}
 \label{fig:2}
\end{figure}


\begin{table}[]
\resizebox{\columnwidth}{!}{%
\tiny
\begin{tabular}{@{}lllll@{}}
\toprule
Dataset  & REKnow  &  w/o KG & w/o Text \\ \midrule
NYT                        &  93.6 (0.17)    & 93.2 (0.16)      &  89.1 (0.16)  \\
WebNLG                     &  89.1 (0.24)    & 88.3 (0.49)     &   67.8  (0.32)    \\
ACE2005                    &  69.4 (0.54)    & 69.2 (0.80)    &    9.2 (0.33)      \\ \bottomrule
\end{tabular}
}
\caption{Analysis of knowledge enhancement. ``REKnow'' is the proposed method.  ``REKnow w/o KG'' means raw text is used. ``REKnow w/o Text'' denotes only found KG information is used.  The F1 scores and standard deviations over 5 seeds are reported.}
\label{table:KG_ablation}
\end{table}


\begin{table}[]
\centering
\resizebox{0.7\columnwidth}{!}{%
\tiny
\begin{tabular}{@{}llll@{}}
\toprule
\multirow{2}{*}{Dataset} & \multicolumn{3}{l}{Found Ext. Info Ratio} \\ \cmidrule(l){2-4} 
                       & Train         & Val         & Test        \\ \midrule
NYT10                  & 1.75          & 1.76        & 1.76        \\
WebNLG                 & 0.93          & 0.91        & 0.97        \\
ACE2005                & 0.75          & 0.71        & 0.66        \\ \bottomrule
\end{tabular}
}
\caption{Statistics of entity linking. ``Found Ext. Info Ratio'' represents the ratio between the size of entities in the text and the size of found external information in entity linking step. Low ratio represents few external information is found and vice versa.}
\label{table: found_el_ratio}
\end{table}

For Relation Classification tasks, we summarize the result in Table \ref{table:main_1}. 
More introduction about the competing methods can be found in ``Related Work'' Section.
``REKnow'' denotes our proposed method.
For TACRED dataset, our generative framework is close to the state-of-the-art method RECENT \cite{Lyu2021RelationCW} method. It shows the effectiveness of our unified framework.
Since the entity types are already provided in TACRED, we replace the grounded entity information with the original entity type in our framework. The result using the original entity type is 74.3 which is lower than the result of the grounded entity information 75.1.
The reason is that original entity type in TACRED is a coarse-grained level entity type while WikiData can provide a fine-grained entity information, which can boost the performance.
For SemEval2010, our method achieves performance comparable to one of the most popular methods BERT-MTB \cite{soares2019matching}.

For Joint Entity and Relation Extraction tasks,  we summarize the result in Table \ref{table:main_2}. 
For NYT dataset, we can find that our ``REKnow'' method outperforms the current SOTA generative-based model, for example, a 0.2 F1 score increase compared with the REBEL method, which is pretrained on a constructed relation dataset ``CROCODILE'' \cite{cabot2021rebel}.
The performance of our method is 2.4 F1 higher than the SOTA extractive-based method Tplinker \cite{wang2020tplinker}.
For ACE2005, ``REKnow'' achieves F1 score with 69.4 (0.54), which is slightly higher than the SOTA method Pure-Alb \cite{zhong2020frustratingly} which is a pipelined framework constructed for entity and relation extraction task. 

Also, from Table \ref{table:main_2}, we notice that ``REKnow'' has the highest recall, which means the method can cover more potential triples. 
The performance of different triple sizes can be found in Figure \ref{fig:2}.
We can see that for NYT and WebNLG datasets, our method has a better performance when the triple size is greater than 1.

\begin{table}[]
\resizebox{1\columnwidth}{!}{%
\begin{tabular}{@{}llllll@{}}
\toprule
Model           & NYT  & WebNLG & ACE2005 & TACRED & Semeval \\ \midrule
T5-large   & 93.6 & 89.1   & 69.4    & 75.1   & 89.8    \\
T5-base   & 92.1 & 87.5   & 64.6    & 71.8   & 89.0   \\
BART-large & 87.3 & 77.8   & 56.5    & 61.1   & 88.4   \\ \bottomrule
\end{tabular}
}
\caption{Pre-trained model comparisons in proposed REKnow method. T5-large model has better performance. }
\label{table:pretrain_model_comparison}
\end{table}
\paragraph{Analysis of Knowledge Enhancement}
In REKnow, both text and entity-related information from KG are used. 
To study the usefulness of external information, we re-train the model with different information source components. 
We summarize the results in Table \ref{table:KG_ablation}. 
To be noticed that the external information in NYT dataset is very useful and achieves 89.1 F1 score without original input text. 
In NYT dataset, adding external KG information can boost the performance from 93.2 (0.16) to 93.6 (0.17), increment is higher than the standard deviation of 5 repetitions.
In WebNLG, similar increment can be observed that 0.9 F1 score increase.
However, in the case of ACE2005, because many entities in ACE2005 are pronoun (he, she, etc). The external information about pronoun cannot contribute much to the relation extraction task. Therefore, there is not significant improvement by incorporating the external knowledge.

To investigate the performance of entity linking, we also report the ``found ext. info ratio'' by calculating the ratio between the size of entities in the text and the size of found external information.
The results for different benchmarks are summarized in Table \ref{table: found_el_ratio}. 
For the NYT10 dataset, the size of grounded entities is much larger than the size of actual entities.
It means the entities widely exist in the plain text, and the many relations among entities are not included as triples in the dataset. It can help explain that, in Table \ref{table:KG_ablation}, ``REKnow w/o Text" can achieve such good performance from the plenty of grounded entity information.


\begin{table}[]
\resizebox{1\columnwidth}{!}{%
\begin{tabular}{@{}lp{7cm}@{}}
\toprule
Error reason     & The truth is incomplete                                                                                                                                                                              \\ \midrule
Dataset          & WebNLG (traceID : test\_578)                                                                                                                                                                         \\ \midrule
Original Text    & 1634: The Bavarian Crisis is the sequel to The Grantville Gazettes .                                                                                                                                \\ \midrule
Text with EL     & 1634: The Bavarian Crisis is the sequel to The Grantville Gazettes .   [gr] 1634: The Bavarian Crisis is an instance of literary work [gr] The   Grantville Gazette is an instance of literary work \\ \midrule
Generated Triple & 1634: The Bavarian crisis precededBy The Grantville Gazettes                                                                                                                                        \\ \midrule
Truth (s, r, o)  & (``1634 : The Bavarian'', ``precededBy'', ``The Grantville Gazettes'')                                                                                                                                   \\ \midrule
Analysis         & ``1634: The Bavarian crisis'' should be treated as a whole.   \\ \midrule \midrule

Error reason     & Generative model has much wider searching space                                                                                                                                                                             \\ \midrule
Dataset          & Webnlg (traceID : test\_633)                                                                                                                                                                     \\ \midrule
Original Text    & Bill Oddie 's daughter is Kate Hardie .                                                                                                                                \\ \midrule
Text with EL     & Bill Oddie 's daughter is Kate Hardie . [gr] Bill Oddie is an instance of   human [gr] Kate Hardie is an instance of human                           \\ \midrule
Generated Triple & Bill Oddie daughter Kate Hardie                                                                                                                                \\ \midrule
Truth (s, r, o)  & (``Bill Oddie'', ``child'', ``Kate Hardie'')                                                                                                                                     \\ \midrule
Analysis         & ``daughter'' is not in the pre-defined relation set     \\ 

\bottomrule
\end{tabular}
}
\caption{Example of Error Analysis. More examples can be found in Table \ref{table:error_full} in the Appendix \ref{appen:B}}
\label{table:error}
\end{table}
\paragraph{Ablation Study}
We study the comparison of whether to use the data augmentation component for entity position-absent case.
In NYT dataset, removing data augmentation will decrease the F1 score 2.0. In WebNLG and ACE2005, no data augmentation step will cause the F1 score drops of 9.8 and 6.1, respectively. 
It shows that exposure bias will affect the performance for generative-based method and using data-augmentation strategy can alleviate the exposure bias significantly. 

We also conduct an ablation study of the pretrained model, BART and T5. The comparison result is summarized in Table \ref{table:pretrain_model_comparison}. 
From the table, we can observe that T5 is generally better than the BART-large model. 
To be specific, in datasets NYT and Semeval, the performances between T5 and BART are close. As for the rest datasets, the BART model decreases about 8 to 10 F1 scores compared with T5.
It would be interesting to study the reason for the heterogeneous difference among datasets. 
We leave that as a direction for future investigation.

In the appendix, one ablation study of hyper-parameter is presented in Table \ref{table:epoch}. 




\paragraph{Error Analysis}

Based on the error analysis in Table \ref{table:error}, we notice that the common error reasons include incomplete truth labels.
The triple provided in the dataset is (``1634: The Bavarian'', ``precededBy'', ``The Grantville Gazettes''), but ``1634: The Bavarian crisis'' is a book title --- REKnow correctly predicted this as a whole entity.
Another error is that generative model may create relation out of the pre-defined relation set. In the second error case in Table \ref{table:error}, ``daughter'' is not a pre-defined relation. 
More cases of error analysis is presented in Table \ref{table:error_full}.
This issue may be addressed by constraint generation, which needs future investigation.

\section{Conclusion}

We proposed a knowledge-enhanced unified generative model for a relational fact extraction task that generates triples sequentially with one pass.
We showed that it outperforms previous state-of-the-art models
on multiple relation extraction benchmarks.

\bibliography{anthology,custom}
\bibliographystyle{acl_natbib}

\clearpage
\appendix
\noindent \Large{\textbf{Appendix:}}
\normalsize
\setcounter{table}{0}
\renewcommand{\thetable}{A\arabic{table}}





\section{Hyper-parameter in Model Implemetation}\label{appen:c}

We adopt the following setting for model training:
\begin{itemize}
    \item learning\_rate : 8e-5 
    \item max\_source\_length : 1024
    \item max\_target\_length : 128
    \item num\_train\_epochs : 10
    \item per\_device\_train\_batch\_size : 4 (for T5-large), 16 (for T5-base, T5-small and BART)
    \item nproc\_per\_node : 8
    \item optimizer : AdamW
    \item lr\_scheduler\_type : linear
\end{itemize}

\paragraph{Ablation study of num\_train\_epochs}:
\begin{table}[]
\resizebox{1\columnwidth}{!}{%
\begin{tabular}{@{}clllll@{}}
\toprule
Epoch No.           & NYT  & WebNLG & ACE2005 & TACRED & Semeval \\ \midrule
3  & 92.6 & 87.0   & 68.3    & 74.6   & 89.1    \\
5  & 92.9 & 87.8   & 68.5    & 76.4   & 89.6    \\ 
7  &  93.5 & 88.6   & 68.4    & 75.4   & 89.2    \\ 
10 &  93.6 & 89.1   & 69.4    & 75.1   & 89.8    \\ \bottomrule
\end{tabular}
}
\caption{Ablation Study of training epoch for REKnow. }
\label{table:epoch}
\end{table}
From the table \ref{table:epoch}, we can find that except TACRED, training 10 epochs can boost the performance. In the main content, we adopt num\_train\_epochs = 10 for all experiments.

\section{Error Analysis}\label{appen:B}

Based on the error analysis in Table \ref{table:error_full}, we find the common error reasons include incomplete truth labels, too wide generation space and missing entity linking result.
In Case 2, the truth triplets should contain all possible relations between a borough ``Manhattan'' and some places including ``Washington Heights'', ``The Battery'' and ``Harlem River''.
In Case 3, if an inversable relation, such as ``org\_alternate\_names", exists, the generated model may generate the triplet in a wrong order compared with golden label.
In Case 4, generative model may create relation out of the pre-defined relation set. This issue may be addressed by constraint generation, which is under future investigation.
In Case 5, the entity linking result may fail to provide enough background information. 
Hence the generated triple confuse with the relation between ``sounds'' and ``species''.

\begin{table*}
\resizebox{2\columnwidth}{!}{%
\begin{tabular}{@{}llp{12cm}@{}}
\toprule
\multicolumn{2}{l}{Error   reason :}                                & The truth is incomplete                                                                                                                                                                                                                                                                                                                                                                                                                                                                                        \\ \midrule
\multicolumn{1}{l|}{Case 1} & \multicolumn{1}{l|}{Dataset}          & Webnlg (traceID : test\_578)                                                                                                                                                                                                                                                                                                                                                                                                                                                                                   \\ \cmidrule(l){2-3} 
\multicolumn{1}{l|}{}       & \multicolumn{1}{l|}{Text with EL}     & 1634 : The Bavarian crisis is the sequel to The Grantville Gazettes .   [gr] 1634: The Bavarian Crisis is an instance of literary work [gr] The   Grantville Gazette is an instance of literary work                                                                                                                                                                                                                                                                                                           \\ \cmidrule(l){2-3} 
\multicolumn{1}{l|}{}       & \multicolumn{1}{l|}{Generated Triple} & 1634 : The Bavarian crisis precededBy The Grantville Gazettes                                                                                                                                                                                                                                                                                                                                                                                                                                                  \\ \cmidrule(l){2-3} 
\multicolumn{1}{l|}{}       & \multicolumn{1}{l|}{Truth (s, r, o)}  & (``1634 : The Bavarian'', ``precededBy'', ``The Grantville Gazettes'')                                                                                                                                                                                                                                                                                                                                                                                                                                               \\ \cmidrule(l){2-3} 
\multicolumn{1}{l|}{}       & \multicolumn{1}{l|}{Analysis}         & ``1634 : The Bavarian crisis'' should be treated as a whole part                                                                                                                                                                                                                                                                                                                                                                                                                                                 \\ \midrule
\multicolumn{1}{l|}{Case 2} & \multicolumn{1}{l|}{Dataset}          & NYT (traceID : test\_582)                                                                                                                                                                                                                                                                                                                                                                                                                                                                                      \\ \cmidrule(l){2-3} 
\multicolumn{1}{l|}{}       & \multicolumn{1}{l|}{Text with EL}     & Through a series of leisurely walks around the island -- from the Battery   to Washington Heights , and from Wall Street to the Harlem River -- Lopate   ruminates on Manhattan 's history , architecture and inhabitants . [gr] The   Battery (Manhattan) is an instance of urban park [gr] Washington Heights,   Manhattan is an instance of neighborhood [gr] Wall Street is an instance of   street [gr] Harlem River is an instance of strait [gr] Manhattan is an   instance of borough of New York City \\ \cmidrule(l){2-3} 
\multicolumn{1}{l|}{}       & \multicolumn{1}{l|}{Generated Triple} & Washington Heights neighborhood\_of Manhattan ; Harlem River   neighborhood\_of Manhattan ; Manhattan contains Washington Heights ; Manhattan   contains Harlem River ; Manhattan contains the Battery;                                                                                                                                                                                                                                                                                                        \\ \cmidrule(l){2-3} 
\multicolumn{1}{l|}{}       & \multicolumn{1}{l|}{Truth (s, r, o)}  & (``Manhattan'', ``contains'', ``Washington Heights''), (``Washington Heights'',   ``neighborhood\_of'', ``Manhattan'')                                                                                                                                                                                                                                                                                                                                                                                                     \\ \cmidrule(l){2-3} 
\multicolumn{1}{l|}{}       & \multicolumn{1}{l|}{Analysis}         & Relations between ``Manhattan'' and ``The Battery'' or   ``Harlem River'' should be included in the truth                                                                                                                                                                                                                                                                                                                                                                                                            \\ \midrule
\multicolumn{1}{l|}{Case 3} & \multicolumn{1}{l|}{Dataset}          & TACRED (traceID : test\_1013)                                                                                                                                                                                                                                                                                                                                                                                                                                                                                  \\ \cmidrule(l){2-3} 
\multicolumn{1}{l|}{}       & \multicolumn{1}{l|}{Text with EL}     & The two companies were preparing to announce that AIG had agreed to sell   [ms] American Life Insurance Co [gr] business , better known as [ms] Alico   [gr] business , for 68 billion dollars in cash and 87 billion in MetLife   equity , the report said .                                                                                                                                                                                                                                                  \\ \cmidrule(l){2-3} 
\multicolumn{1}{l|}{}       & \multicolumn{1}{l|}{Generated Triple} & American Life Insurance Co org\_alternate\_names Alico ;                                                                                                                                                                                                                                                                                                                                                                                                                                                       \\ \cmidrule(l){2-3} 
\multicolumn{1}{l|}{}       & \multicolumn{1}{l|}{Truth (s, r, o)}  & (``Alico'', ``org\_alternate\_names'', ``American Life Insurance Co'')                                                                                                                                                                                                                                                                                                                                                                                                                                               \\ \cmidrule(l){2-3} 
\multicolumn{1}{l|}{}       & \multicolumn{1}{l|}{Analysis}         & gnerative model is not sensitive to inverseable triple like : org's   alternate name                                                                                                                                                                                                                                                                                                                                                                                                                           \\ \midrule
\multicolumn{2}{l}{Error reason :}                                  & Generative model has too wider searching space                                                                                                                                                                                                                                                                                                                                                                                                                                                                 \\ \midrule
\multicolumn{1}{l|}{Case 4} & \multicolumn{1}{l|}{Dataset}          & Webnlg (traceID : test\_633)                                                                                                                                                                                                                                                                                                                                                                                                                                                                                   \\ \cmidrule(l){2-3} 
\multicolumn{1}{l|}{}       & \multicolumn{1}{l|}{Text with EL}     & Bill Oddie 's daughter is Kate Hardie . [gr] Bill Oddie is an instance of   human [gr] Kate Hardie is an instance of human                                                                                                                                                                                                                                                                                                                                                                                     \\ \cmidrule(l){2-3} 
\multicolumn{1}{l|}{}       & \multicolumn{1}{l|}{Generated Triple} & Bill Oddie daughter Kate Hardie                                                                                                                                                                                                                                                                                                                                                                                                                                                                                \\ \cmidrule(l){2-3} 
\multicolumn{1}{l|}{}       & \multicolumn{1}{l|}{Truth (s, r, o)}  & (``Bill Oddie'', ``child'', ``Kate Hardie'')                                                                                                                                                                                                                                                                                                                                                                                                                                                                         \\ \cmidrule(l){2-3} 
\multicolumn{1}{l|}{}       & \multicolumn{1}{l|}{Analysis}         & ``daughter'' is not in the pre-defined relation set                                                                                                                                                                                                                                                                                                                                                                                                                                                              \\ \midrule
\multicolumn{2}{l}{Error reason :}                                  & Entity linking is missing                                                                                                                                                                                                                                                                                                                                                                                                                                                                                      \\ \midrule
\multicolumn{1}{l|}{Case 5} & \multicolumn{1}{l|}{Dataset}          & semeval (traceID : test\_876)                                                                                                                                                                                                                                                                                                                                                                                                                                                                                  \\ \cmidrule(l){2-3} 
\multicolumn{1}{l|}{}       & \multicolumn{1}{l|}{Text with EL}     & the word ``song '' is used to describe the pattern of regular and   predictable [ms] sounds [gr] NA made by some [ms] species [gr] NA of whales ,   notably the humpback whale .                                                                                                                                                                                                                                                                                                                              \\ \cmidrule(l){2-3} 
\multicolumn{1}{l|}{}       & \multicolumn{1}{l|}{Generated Triple} & species Producer\_Product sounds ; sounds Product\_Producer species ;                                                                                                                                                                                                                                                                                                                                                                                                                                          \\ \cmidrule(l){2-3} 
\multicolumn{1}{l|}{}       & \multicolumn{1}{l|}{Truth (s, r, o)}  & (``sounds'', ``Effect\_Cause'', ``species''), (``species'', ``Cause\_Effect'',   ``sounds'')                                                                                                                                                                                                                                                                                                                                                                                                                               \\ \cmidrule(l){2-3} 
\multicolumn{1}{l|}{}       & \multicolumn{1}{l|}{Analysis}         & The model treat the relation between ``sounds'' and   ``species'' as product-type relation when Entity Linking is missing   in this case                                                                                                                                                                                                                                                                                                                                                                           \\ \bottomrule
\end{tabular}
}
\caption{Error Analysis for REKnow. }
\label{table:error_full}
\end{table*}

\end{document}